%% file: main.tex
\documentclass[sigconf,bookmarksnumbered,unicode,colorlinks,linkcolor=blue]{acmart}

\acmSubmissionID{471}

\acmJournal{TOG}
\acmVolume{41}
\acmNumber{4}
\acmYear{2022}
\acmMonth{8}

\usepackage{multirow}
\usepackage{algorithm}
\usepackage{algorithmic}

\setcitestyle{nosort,square}

\microtypesetup{protrusion=true,factor=750,expansion=true,stretch=10,shrink=5,final}
\newcommand{\rev}[1]{{\color{black}{#1}}}
\newcommand{\finalrev}[1]{{\color{black}{#1}}}

\copyrightyear{2022}
\acmYear{2022}
\setcopyright{rightsretained}
\acmConference[SA '22 Conference Papers]{SIGGRAPH Asia 2022 Conference
Papers}{December 6--9, 2022}{Daegu, Republic of Korea}
\acmBooktitle{SIGGRAPH Asia 2022 Conference Papers (SA '22 Conference
Papers), December 6--9, 2022, Daegu, Republic of
Korea}\acmDOI{10.1145/3550469.3555413}
\acmISBN{978-1-4503-9470-3/22/12}

\begin{document}

\title{Differentiable Point-Based Radiance Fields\\
for Efficient View Synthesis}


\settopmatter{authorsperrow=4}
\author{Qiang Zhang}
\email{qz9238@princeton.edu}
\orcid{0000-0002-4483-1039}
\affiliation{%
  \country{Princeton University, USA}
}

\author{Seung-Hwan Baek}
\email{sb38@princeton.edur}
\orcid{0000-0002-2784-4241}
\affiliation{%
   \country{Princeton University, USA}
}

\author{Szymon Rusinkiewicz}
\email{smr@princeton.edu}
\orcid{0000-0002-4253-2588}
\affiliation{%
   \country{Princeton University, USA}
}

\author{Felix Heide}
\email{fheide@princeton.edu}
\orcid{0000-0002-8054-9823}
\affiliation{%
   \country{Princeton University, USA}
}

\renewcommand\shortauthors{Zhang et al.}

\begin{abstract}
			We propose a differentiable rendering algorithm for efficient novel view synthesis. By departing from volume-based representations in favor of a learned point representation, we improve on existing methods more than an order of magnitude in memory and runtime, both in training and inference.
			The method begins with a uniformly-sampled random point cloud and learns per-point position and view-dependent appearance, using a differentiable splat-based renderer to train the model to reproduce a set of input training images with the given pose.
			Our method is up to 300 $\times$ faster than NeRF in both training and inference, with only a marginal sacrifice in quality, while using less than 10~MB of memory for a static scene.
			For dynamic scenes, our method trains two orders of magnitude faster than STNeRF and renders at a near interactive rate, while maintaining high image quality and temporal coherence even without imposing any temporal-coherency regularizers. 
\end{abstract}


\begin{teaserfigure}
\vspace{-3mm}
\centering
\includegraphics[width=\textwidth]{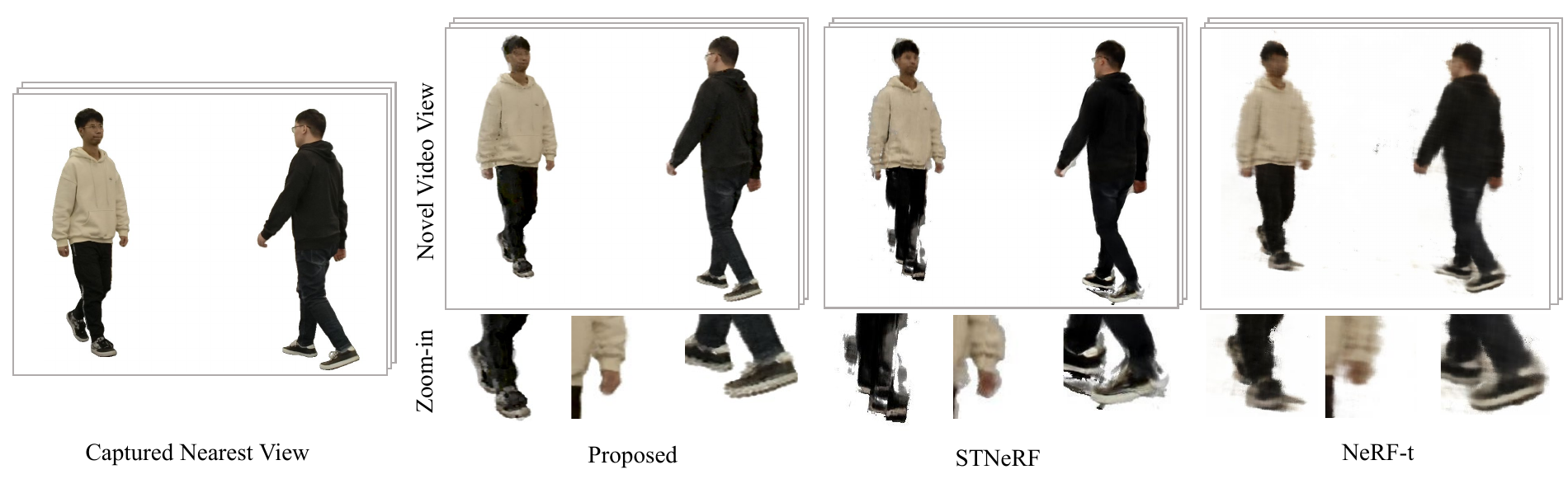}\vspace{-0.1cm}
 \flushright{
\begin{tabular}{rp{0.2cm}rp{3.5cm}rp{3.75cm}rp{2.15cm}}
 &  &
\textbf{Training :} & 30 min &
 & 50 h \,\, \textbf{(100 $\times$)} &
 & 100 h \,\, \textbf{(200 $\times$)} \\
 &  & 
\textbf{Inference :} & 40 ms &  
& 30 s \,\,\, \textbf{(750 $\times$)}&  
 & 26 s \,\,\,\,\; \textbf{(650 $\times$)}\\
 &  & 
\textbf{PSNR :} & 34.6 dB &  
& 32.1 dB &  
 & 28.9 dB \\
\end{tabular}
}
\vspace{-2.5mm}
\label{fig:teaser}
\caption{Novel Video View Synthesis with Differentiable Point-Based Radiance Fields. We introduce a \rev{scene representation that models radiance} with a point cloud augmented with local spherical harmonic coefficients per point. This representation is not only memory efficient but also two orders of magnitude faster in training and inference than the coordinate-MLP based approach that relies on ray-marching, which mandates evaluating a learned network per ray sample. In contrast, the proposed method allows for efficient view synthesis via differentiable splat rendering. We test our method on novel view and novel video view synthesis tasks, where it outperforms existing methods, including \rev{recent NeRF-based approaches}~\cite{zhang2021editable,mildenhall2020nerf} shown here for \rev{multi-view video data}~\cite{zhang2021editable}, in both quality and runtime.} 
\end{teaserfigure}

%
%

	\ccsdesc[500]{Computing methodologies~Rendering}
	
%
%

\keywords{Neural Rendering, Image-based Rendering, Novel View Synthesis}

\maketitle


\input{defs}


\section{Introduction}
\label{sec:intro}
\input{1_introduction}

\section{Related Work}
\label{sec:related}
\input{2_related}

\section{Differentiable Point Radiance Fields}
\label{sec:math}
\input{3_math}



\section{Assessment}
\label{sec:assessment}
\input{5_results}

\section{Conclusion}
\label{sec:conclusion}
\input{8_conclusion}


\bibliographystyle{ACM-Reference-Format}
\bibliography{references}

\newpage
\rev{\section*{Appendix}}
\input{appendix}
\end{document}

%% file: defs.tex
\newcommand{\vect}[1]{\mathbf{#1}}
\newcommand{\mat}[1]{\mathbf{#1}}

\newcommand\note[1]{\textcolor{red}{#1}}

\newcommand{\psf}{\rho}

\newcommand{\argmin}[1]{\stackrel[\{ #1 \}]{}{\textrm{arg min}}}
\newcommand{\minimize}[1]{\stackrel[\{ #1 \}]{}{\textrm{minimize}}}

%% file: 1_introduction.tex
Neural rendering has emerged as an effective tool to model real-world scenes~\cite{mildenhall2020nerf}, synthesize novel views~\cite{lombardi2019neural}, and advance single-shot learning~\cite{yu2021pixelnerf}. Specifically, existing work has made it possible to generate photorealistic novel views at unseen viewpoints from unorganized multi-view training images.
Synthesizing novel views could not only enable after-shot visual effects in consumer photography and virtual reality but also implicitly/explicitly entails the geometric understanding of a scene.
Utilizing an efficient view-synthesis network in an analysis-by-synthesis framework may further allow these techniques to become the missing link for inverse-rendering applications in diverse application domains.

Neural radiance fields (NeRF)~\cite{mildenhall2020nerf} have spearheaded neural rendering by modeling a 5D radiance field in a memory-efficient manner with an implicit scene representation in the form of a multi-layer perceptron (MLP).
A large body of work has built on this method using implicit scene representations for view synthesis, investigating generalizability~\cite{yu2021pixelnerf} and rendering quality~\cite{Wizadwongsa2021NeX}.
However, a drawback of modern neural rendering methods is their computational efficiency.
As NeRF relies on volumetric rendering, training and inference of a model requires many evaluations and memory accesses of the radiance fields: typically more than 100 evaluations for each ray.
As a result, for medium-sized static scenes, NeRF takes over ten hours of training and 30 seconds of inference for each scene on a single GPU. 
The limited computational efficiency of NeRF-based methods prevents the direct applicability to video applications, where fast training and inference at a low memory footprint are mandated.
Recent methods have attempted to tackle this problem by pretraining a multi-view stereo model on a large-scale dataset and fine-tuning it for test inputs~\cite{chen2021mvsnerf}, or using an explicit radiance volume~\cite{yu2021plenoctrees}.
While these methods outperform NeRF in efficiency, they are not capable of accelerating both training and inference speed and require a large pretraining dataset as well as significant storage.

In this work we explore an alternative approach for retaining the highly-desirable end-to-end differentiability of NeRF, while utilizing an alternate radiance field representation.  Our insight is that, even though the MLP used by NeRF is compact and relatively efficient to evaluate once, using it for image synthesis is fundamentally a ``backward mapping'' operation that requires multiple evaluations of radiance along a ray.  Shifting to a representation that allows ``forward mapping,'' i.e., creating an image via a single forward pass over the representation of the radiance field, enables modest gains in storage but, more importantly, efficiency gains of multiple orders of magnitude at both training and rendering time.

Specifically, we use a representation for radiance based on RGB spherical harmonic coefficients stored at points in 3D. Image synthesis is then accomplished via a differentiable splat-based rasterizer that operates as a single forward pass. \rev{The rendering pipeline from point primitives and spherical harmonic coefficients to images is end-to-end differentiable, allowing for optimizing the position and appearance attributes with a coarse-to-fine first-order optimization method.} 

Our strategy yields several benefits: smaller storage, faster training, and more efficient rendering that avoids volumetric ray marching.  The cost is reduced flexibility in radiance representation, but this is not always a drawback.  Indeed, with few input views, the regularization provided by our method leads to higher reconstruction quality.  However, when many input views are provided, our method results in slightly lower quality (for a fixed spherical harmonic degree). \rev{Our approach learns point clouds with position and appearance without a neural network or separate geometry estimation (e.g., via COLMAP) step. As such, it allows for efficient inference \emph{and} training.}

Specifically, we make the following contributions in this work:
\begin{itemize}
\setlength\itemsep{0.5em}
    \item We introduce a point-based radiance representation that, together with differentiable splatting, is 300~$\times$ faster than neural volumetric methods for view synthesis in both training and inference. 
    
    \item We analyze the proposed point-cloud representation in terms of memory, inference and training run-time, and show that it outperforms state-of-the-art neural rendering algorithms~\cite{yu2021plenoxels} in compute while being two orders of magnitude smaller in memory.
    
    \item We demonstrate that our method is efficient enough in memory, training time, and rendering time to enable novel-view synthesis for video, even without establishing explicit frame-to-frame connections.  We validate that the proposed method outperforms existing methods in quality, run time, and memory footprint.
\end{itemize}
\vspace{-1mm}
\rev{
We have released all code and data required to reproduce the findings from this work. Here is the \href{https://github.com/sjtuzq/point-radiance}{link}. 
}
\vspace{-1mm}



\paragraph{Limitations}
The proposed method requires an object mask to generate the initial point cloud, and thus can only render the foreground scene components. This limitation may be potentially lifted by replacing the hard multi-view consistency function with a soft one: setting a threshold to filter based on projected pixel variance. As popular existing multi-view and video datasets contain masks, we leave such a potential extension for future work. \rev{The proposed representation which relies on splatting can also not handle semi-transparent objects, e.g., objects with fur. 
}
\vspace{-2mm}

%% file: 2_related.tex
\paragraph{Neural Rendering}
A rapidly growing body of work investigates introducing differentiable modules into rendering approaches, turning conventional rendering methods into powerful \rev{inverse} tools for capturing, reconstructing, and synthesizing real-world scenes. Among many rendering principles, volumetric rendering has been \rev{actively} studied in conjunction with neural networks~\cite{mildenhall2020nerf} thanks to its learning-friendly structure. In this context, various \rev{network} architectures have been proposed to model scene radiance, including convolutional neural network (CNN) in \rev{Neural Volumes}~\cite{lombardi2019neural} and multi-layer perceptron (MLP) in NeRF~\cite{mildenhall2020nerf}.
While the synthesized images from these methods achieve unprecedented fidelity, one of the drawbacks is their generalization ability. To tackle this challenge, image features \rev{can be} explicitly exploited either from a single image~\cite{yu2021pixelnerf} or multiple images~\cite{wang2021ibrnet,chen2021mvsnerf,liu2021neural}. 
Another line of neural-rendering method\rev{s} utilizes a multi-plane image (MPI) as a scene representation for high-quality view synthesis~\cite{zhou2018stereo,Wizadwongsa2021NeX}.
Lastly, using point clouds has been studied for neural rendering with iterative filtering~\cite{rosenthal2008image} and multi-scale approaches~\cite{marroquim2007efficient,meshry2019neural,ruckert2021adop}. 
\rev{Among the many successful approaches, NeRF-based volumetric methods in particular have achieved remarkable success.}

\rev{However}, volumetric methods are not without limitations, and one major problem is their computational efficiency. While occupying only a small memory footprint, NeRF-based models typically require hours of training in a modern GPU, and inference takes minutes. 
This low computational efficiency prevents NeRF methods from being used in applications that require fast training and inference.
Recently, researchers devised methods to accelerate inference, at the cost of additional memory incurred: Garbin et al.~\shortcite{garbin2021fastnerf} cache explicit volumes, Yu et al.~\shortcite{yu2021plenoctrees} propose an octree structure, and Reiser et al.~\shortcite{reiser2021kilonerf} replace a large MLP with multiple small-size MLPs. Although these methods can speed up the rendering process, the training is still volumetric, making existing approaches slow. \rev{Recently, Lombardi et al.~\shortcite{lombardi2021mixture} proposed a hybrid scene representation with primitive-based and volumetric components. As such, this method inherits some of the drawbacks of volumetric methods, including training time: this work requires 5 days of training time for a single scene, orders of magnitude more than the proposed method.}
\finalrev{Hedman et al.~\shortcite{hedman2021baking} propose a post-training approach for fast rendering with NeRFs and Neff et al.~\shortcite{neff2021donerf} introduce the use of depth oracle to speed up the model training speed.}
Yu et al.~\shortcite{yu2021plenoxels} use explicit volumes instead of an MLP and directly optimizes the volume parameters. 
However, this results in gigabytes of memory consumption for each scene, hence an inevitable explosion with terabytes of memory for multi-frame video.
In this work, we propose a computationally-efficient rendering method that accelerates both training and inference, achieving 2~$\times$ faster inference, 3~$\times$ faster training, and 100~$\times$ lower memory footprint than Yu et al.~\shortcite{yu2021plenoxels}. 

%
\begin{figure*}[t]
  \centering
   \includegraphics[width=0.95\linewidth]{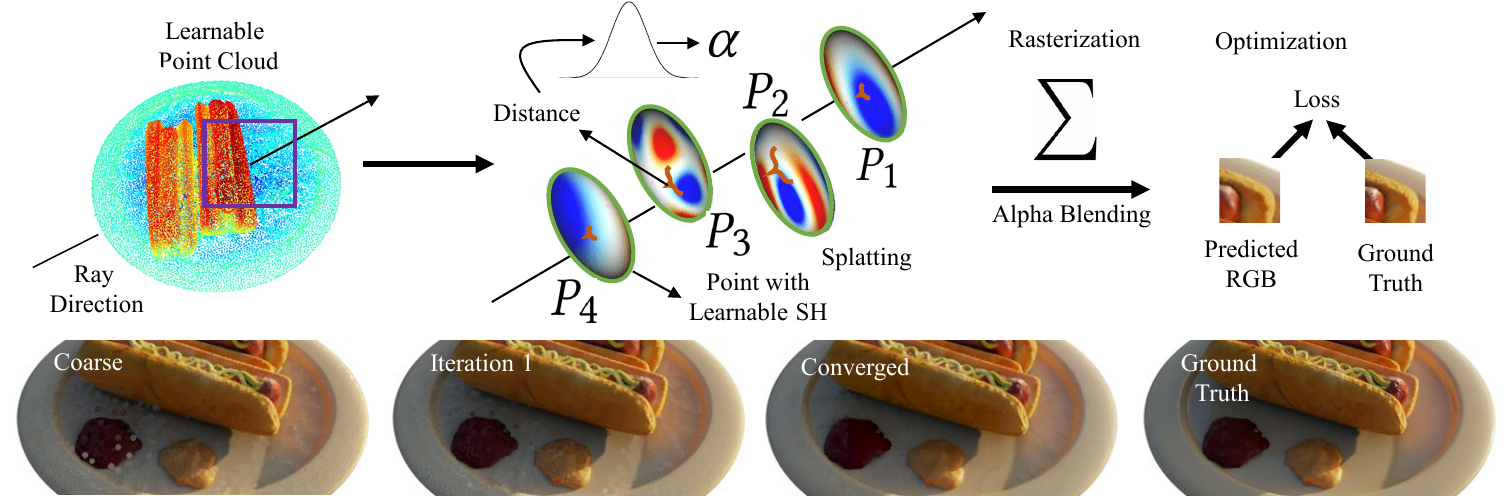}
   \vspace{-4mm}
   \caption{\textbf{Learning Point-Based Radiance Fields.} The proposed method learns a point cloud augmented with RGB spherical harmonic coefficients per point. Image synthesis is accomplished via a differentiable splat-based rasterizer that operates as a single forward pass, without the need for ray-marching and hundreds of evaluations of coordinate-based networks as implicit neural scene representation methods. Specifically, the point rasterizer computes an alpha value from the ray-point distance with a Gaussian radial basis function weight. The learnable radiance and the alpha values are rasterized via alpha blending to render the image. The entire rendering pipeline is differentiable and we optimize for the point position and spherical harmonic coefficients end-to-end. We use a hybrid coarse-to-fine strategy (bottom) to train the entire model and successively refine the geometry represented by the learned point cloud.}
   \label{fig:sh}
  \vspace{-5mm}
\end{figure*}

\vspace{-2mm}
\paragraph{Multi-view Geometry Reconstruction}
Our work falls into the category of multi-view geometry reconstruction methods studied for decades in computer graphics and vision. This direction traces back to popular multi-view stereo methods using local photometric consistency and global visibility constraints~\cite{furukawa2009accurate} and geometric priors~\cite{schonberger2016pixelwise}.
Recently, neural networks have accelerated multi-view geometry acquisition~\cite{yao2018mvsnet,im2019dpsnet,han2020drwr}.
Notably, Han et al.~\shortcite{han2020drwr} use multi-view projection and exploit silhouette images to estimate shape, demonstrating state-of-the-art performance. \rev{Recently, Wiles et al.~\shortcite{wiles2020synsin} (SynSin) and Thonat et al.~\shortcite{thonat2021video} propose single-shot video prediction and mesh-based multi-view video methods with technical approaches orthogonal to ours. Specifically, none of these existing methods can model the geometry and radiance simultaneously. In addition, they typically require a separate dataset to pretrain the model for a geometry prior. The proposed method optimizes a learned point cloud directly for each scene that does not require additional learned or explicit priors. \finalrev{Recent work~\cite{kondo2021vaxnerf} explores how knowledge of the visual hull can help accelerate the neural rendering process, and this method cannot improve inference time}. }

\vspace{-2mm}
\paragraph{View-dependent Modeling and Rendering}
Modeling and rendering view-dependent scene radiance has been studied extensively in graphics and vision.
Surface light fields~\cite{wood2000surface} is based on a view-dependent level-of-detail rendering for meshes with subdivision connectivity. 
Using spherical harmonics in this context has been proposed as an effective way to represent a view-dependent function, with broad adoption in rendering and modeling~\cite{cabral1987bidirectional,sillion1991global,ramamoorthi2001efficient,ramamoorthi2001relationship,basri2003lambertian,sloan2003clustered,sloan2002precomputed}.
Recent works apply view-dependent models to reflectance~\cite{chen2021invertible,sztrajman2021neural} via implicit BRDF modeling and use spherical harmonics to describe view-dependent radiance distribution~\cite{yu2021plenoctrees,Wizadwongsa2021NeX}.
The proposed method also \rev{uses} spherical harmonics to model view-dependence while maintaining a lightweight and efficient model.

\vspace{-2mm}
\paragraph{Point-based Rendering}
Point-based \rev{rendering via splatting} has been explored for decades, offering computational efficiency \rev{as} an elegant scene representation~\cite{pfister2000surfels,zwicker2001surface,zwicker2004perspective,zwicker2002pointshop}. \rev{Recent works explore combining deep learning with surface splatting, e.g., Yifan et al. \shortcite{yifan2019differentiable} propose a differentiable algorithm for point-based surface and normal processing and does not have an appearance model and does not produce radiance outputs. In contrast, \finalrev{the proposed method optimizes for point positions and spherical harmonic coefficients in an end-to-end fashion, and it focuses on high-quality image-based rendering and novel view synthesis.} Aliev et al.~\shortcite{aliev2020neural} propose a point-based rendering method that requires a \emph{given point cloud} aggregated from RGB-D captures, while the proposed method optimizes for point clouds from RGB-only inputs. In contrast to our network-free method, this method also uses a rendering network that requires a day of training time. Lassner et al.~\shortcite{lassner2021pulsar} generate features by rasterization of spheres in a \emph{deferred neural rendering} approach using an image-to-image rendering network (Pix2PixHD). The proposed method synthesizes an RGB image directly by splatting a point cloud with a per-point appearance model. We report in this manuscript that this results in an order of magnitude faster training time while providing better results. Kopanas et al.~\shortcite{kopanas2021point} use rasterized encoded point clouds with features provided by conventional \rev{structure-from-motion methods}, i.e., using COLMAP to build a dense point cloud, and SIBR for the depth and normal maps. A neural network then translates features into images for novel views. The feature encoding and decoding are learned, leading to long training times.  Moreover, their inference stage selects image features from the training set and thus needs to store all these training features, resulting in storage requirements of multiple gigabytes in contrast to less than ten megabytes for the proposed method. Similarly, R{\"u}ckert et al.~\shortcite{ruckert2021adop} assume \emph{known point clouds}, recovered by COLMAP. Given a point cloud, their method employs a deferred neural rendering approach similar to \rev{Lassner et al.~\shortcite{lassner2021pulsar}} but with a multi-scale feature-to-image transfer network. In contrast to the proposed method that requires a few minutes of training time, R{\"u}ckert et al. train for 12 hours.
\finalrev{Very recently, Point-NeRF~\cite{xu2022point} proposes a point-based rendering method, which, in contrast to the proposed method, is not network-free and relies on relatively slow volume-rendering. As such, the rendering time of Point-NeRF is comparable to NeRF, about 400x slower than the proposed method.}
}

%% file: 3_math.tex
\newcommand{\T}{^{\intercal}} 
\newcommand{\Image}[1]{I_{#1}}
\newcommand{\rendered}[1]{\hat{I}_j}
\newcommand{\nimg}{N} 
\newcommand{\Point}[1]{P_{#1}} 
\newcommand{\point}[2]{p_{#1}^{#2}} 
\newcommand{\npts}{n} 
\newcommand{\radiance}[1]{H_{#1}}
\newcommand{\radcoeff}[3]{h_{#1,#2#3}}
\newcommand{\mask}[1]{m_{#1}}
\newcommand{\rot}[1]{R_{#1}}
\newcommand{\trans}[1]{t_{#1}}
\newcommand{\intrinsics}[1]{M_{#1}}
\newcommand{\proj}[1]{\left(#1\right)^\downarrow}
\newcommand{\lmax}{l_\textrm{max}}
\newcommand{\view}[2]{v_{#1}^{#2}}
\newcommand{\col}[2]{c_{#1}^{#2}}
\newcommand{\sph}[2]{Y_{#1}^{#2}}
\newcommand{\pixel}{u}
\newcommand{\opacity}[2]{\alpha_{#1}^{#2}(\pixel)}
\newcommand{\contribution}[2]{A_{#1}^{#2}(\pixel)}
\newcommand{\loss}{\mathcal{L}}


In this section, we describe the proposed differentiable point-based rendering method, which is illustrated in Fig.~\ref{fig:sh}. We assume a set of images $\Image{j}$, $j = 1 \ldots \nimg$ as input, with known camera intrinsics and extrinsics. We learn a model of the scene as a point cloud with a per-point radiance model based on spherical harmonics. Learning minimizes the difference between the rendered scene and the original images, by updating the position and appearance of each point. The use of a differentiable splat-based rasterizer allows for gradient-based updates, while being orders of magnitude more efficient than an implicit volume-based renderer. The point cloud, the proposed method is operating on, is optimized end-to-end in a coarse-to-fine iterative fashion, starting from a rejection-sampled initial point cloud. We describe the components of the proposed method in the following.

\begin{figure}[t!]
  \centering
\includegraphics[width=0.9\linewidth]{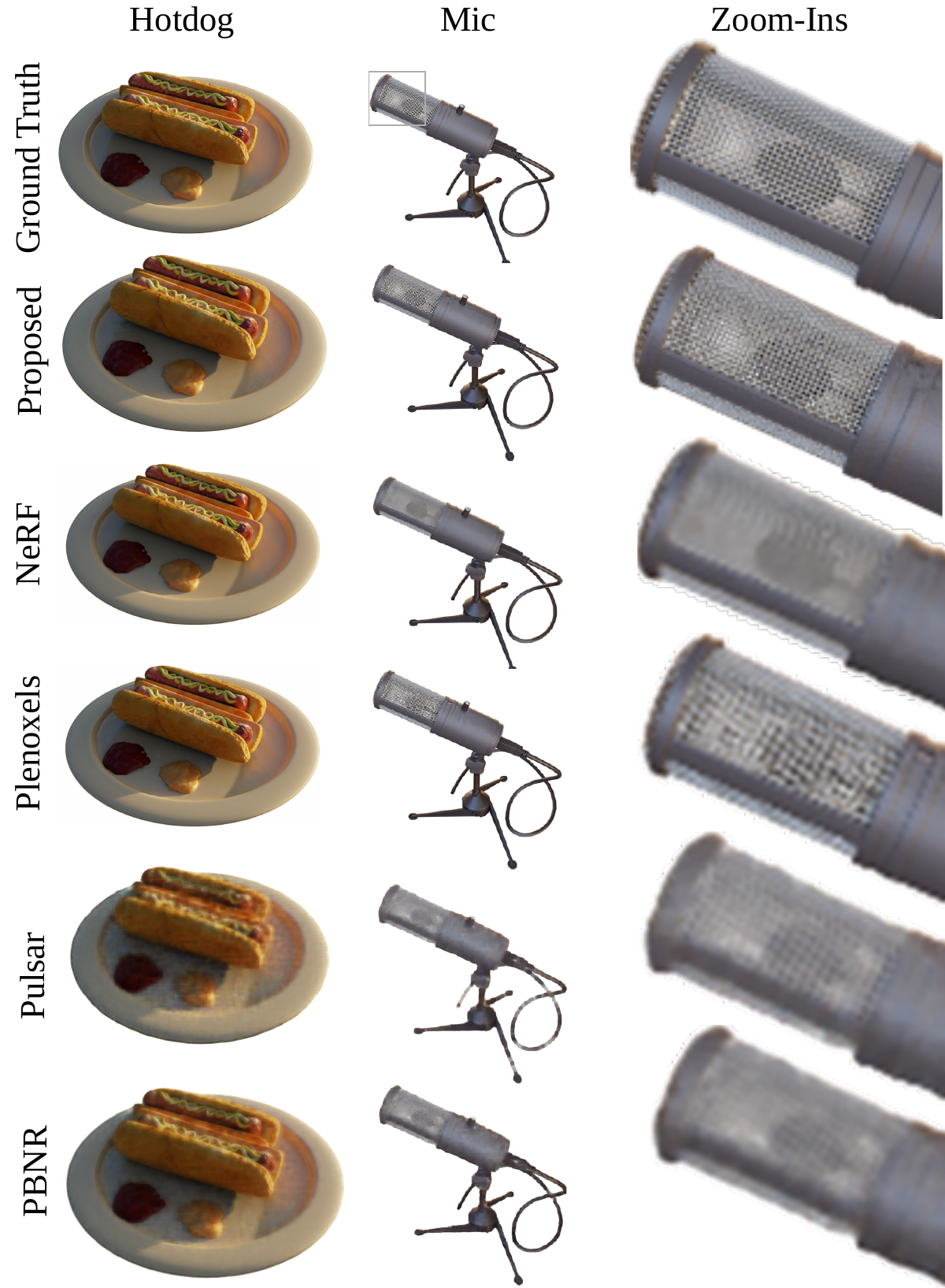}
   \caption{\textbf{Synthetic View Synthesis for Static Scenes.} We represent \rev{two synthetic scenes~\cite{mildenhall2020nerf} with the proposed point-based radiance model}. Trained only in a few minutes on multi-view images, the proposed method achieves rendering quality comparable to state-of-the-art volumetric rendering approaches such as Plenoxels~\cite{yu2021plenoctrees}, while being more than 3~$\times$ faster in training, 2~$\times$ faster in inference and 100~$\times$ less memory-hungry.}
   \label{fig:synthetic}
   \vspace{-6mm}
\end{figure}

\subsection{Point-Based Scene Radiance}
We model the scene as a set of points $\Point{i} = (x_i,y_i,z_i)\T$, $i = 1, \ldots, \npts$, with the parameters $\radiance{i}$ of a view-dependent radiance model per point. These 3D positions and radiance model parameters serve as the unknowns that will be updated by the optimization method to reconstruct the scene.

\paragraph{Geometric Model}
Each 3D point $\Point{i}$ projects to a 2D position $\point{i}{j}$ in each image $\Image{j}$.  To model the projection operation, we assume that we know the extrinsic rigid-body transform $(\rot{j},\trans{j})$ and intrinsic matrix $\intrinsics{j}$ for each camera, and then compute
\begin{equation}\label{eq:projection}
    \point{i}{j} = \proj{\intrinsics{j} (\rot{j}\Point{i} + \trans{j})},
\end{equation}
where $\proj{x,y,z} = (x/z, y/z)$ denotes the perspective divide. \rev{Here $P_i$ are represented as the learnable neural parameters}. Note that visibility is resolved by the rendering stage (see \rev{Section}~\ref{sec:splat}) via depth sorting and alpha blending.

\paragraph{Photometric Model}
We model the view-dependent radiance of each point with spherical harmonics~\cite{cabral1987bidirectional} up to a maximum degree $\lmax$:
\begin{equation}
\radiance{i} = \{ \radcoeff{i}{l}{m} \;|\; 0 \leq l \leq \lmax, \; -l \leq m \leq l\}.
\end{equation}
\rev{Note that $\radcoeff{i}{l}{m}$ here are learnable.} If we then denote the view direction of point $i$ in image $j$ as
\begin{equation}
\view{i}{j} = \frac{\rot{j}\Point{i} + \trans{j}}{\|\rot{j}\Point{i} + \trans{j}\|},
\end{equation}
we can represent the color of the point in that image as
\begin{equation}
 \label{eq:sh}
\col{i}{j} = \sum_{l=0}^{\lmax} \sum_{m=-l}^{l} \radcoeff{i}{l}{m} \, \sph{l}{m} (\view{i}{j}).
\end{equation}



\paragraph{Point Cloud Initialization}
We initialize points randomly within the visual hull defined by the image-space masks.  Specifically, we use a rejection sampling method in which we uniformly sample random 3D points, project them into each image, and keep them only if they fall within all the masks. We initialize the point appearance by sampling the spherical harmonics parameter \rev{from the normal distribution with zero mean and standard deviation of one}.

Our method is thus dependent on the availability and quality of the initial masks.  However, we found that the optimization still produces good results if the masks are inaccurate or there are concavities not modeled by the visual hull. 
\rev{Further details on initialization are provided in the Appendix.}


\subsection{Differentiable Splat Rendering\label{sec:splat}}
Given a point cloud model, we render an image via point splatting, see \rev{Zwicker et al.~\shortcite{zwicker2001surface}} for background on splatting methods. Specifically, we use an image-space radial basis function (RBF) kernel to represent the contribution of each point $\point{i}{j}$ to each pixel $\pixel$ in image $j$. We employ a Gaussian kernel of radius $r$, computing the opacity as
\begin{equation}
\opacity{i}{j} = \frac{1}{\sqrt{2\pi r^2}} \, e^{-\frac{\|\point{i}{j} - \pixel\|^2}{2 r^2}}.
\end{equation}
 
After obtaining the $\alpha$ map, we sort the points by $z$-distance and compute the final rendered image $\rendered{j}$ via alpha blending from front to back. That is, we compute
\begin{equation}
\rendered{j}(\pixel) = \sum_{i=1}^\npts \contribution{i}{j} \, \col{i}{j},
\end{equation}
where $\contribution{i}{j}$ represents the net contribution of the $i$-th point:
\begin{equation}
\contribution{i}{j} = \opacity{i}{j} \, \prod \limits_{k=1}^{i-1}\left(1-\opacity{k}{j}\right),
\end{equation}
where the points are taken as sorted front-to-back and $\col{i}{j}$ is the per-point color computed via Eq.~\ref{eq:sh}. Note that, as an optimization, we only consider points within a certain maximum distance of $\pixel$ with non-negligible opacity.

\subsection{Point Cloud Training} \label{sec:training}
To train a point cloud model that predicts the image $\Image{j}$, we employ a loss function consisting of an MSE and total variation loss, that is
\begin{equation}
\loss = \sum_{j=1}^{\nimg} ||\Image{j}-\rendered{j}||_2^2 + \lambda TV(\rendered{j}).
\end{equation}
\rev{The total variation term represents the conventional anisotropic total variation penalty $TV(\cdot) = |\nabla_x \cdot|_1 + |\nabla_y \cdot|_1$, that is the $\ell_1$-norm of the predicted image gradients. } Since the loss function, splatting rasterizer, and point cloud geometric and photometric models are all fully differentiable, we are able to learn the point positions and their parameters end-to-end using stochastic gradient descent optimization.

\rev{To train the method we propose a non-synchronous learning strategy for optimizing the point position and harmonics parameters. Specifically, the initial learning rate for the spherical harmonics radiance model and the point positions are $3e-3$ and $8e-4$, respectively. Both of them are exponentially decayed for every epoch (20 epochs in total) with a factor of $0.93$. The TV weight $\lambda$ is set to $0.01$. The splatting radius $r$ and the rasterization $n$ are set as $0.008$ and $15$ respectively. }
We use $\lmax = 2$ for all of the results in the paper, since in our experiments using a higher degree resulted in a minimal quality gain for significant extra storage and training time. Section~\ref{sec:ablation} contains an ablation study comparing spherical harmonics to just a single color, which effectively corresponds to $\lmax = 0$.

\paragraph{Hybrid Coarse-to-Fine Updates}
To learn an accurate point cloud representation, we repeat the following several stages of a coarse-to-fine scheme twice, also illustrated in Fig.~\ref{fig:sh}:

\rev{
\begin{enumerate}
    \item{Voxel-based Point Cloud Reduction:} We assume a voxel discretization (without actually voxelizing the scene), and, for each voxel, we aggregate all the points inside that voxel into one single point by averaging the position and the spherical harmonics of these points. The motivation behind this step is to avoid the formation of dense point clusters and improve the representation efficiency. 
    
    \item{Outlier Removal:} For each point, we compute the distance to its neighbors and the standard deviation of these distances. We then filter out the outliers whose standard deviation is beyond a threshold, hence removing outlier points.
    
    \item{Point Generation:} In this stage, we insert new points into the point cloud by generating a single new point for each point in the point cloud. We set all parameters (including position) to the average of the point's nearest $K$ neighbors. 
\end{enumerate}
}

\rev{We refer to Alg.~\ref{alg:hybrid} for this coarse-to-fine update in the Appendix}. As a final step, we fine-tune the model with half of the original learning rate in a refinement step.


\begin{table*}[t!]
\caption{\textbf{Static Novel View Synthesis Evaluation on the Synthetic Blender Dataset.} We evaluate the proposed method and comparable baseline approaches for novel view synthesis on the static Blender scenes from~\cite{mildenhall2020nerf}. \rev{The proposed model does not require an extra dataset for pretraining and improves on existing methods in training, inference speed and model size, at cost of only a small reduction in quality. Specifically, although the concurrent Plenoxels~\shortcite{yu2021plenoxels} achieves better quality, our model is two magnitudes smaller than theirs. We also compare here to the Plenoxels\_s model from~\cite{yu2021plenoxels} (Plenoxels with smaller volume resolution), which achieves worse rendering quality with a larger model size.}}
\vspace{-3mm}
\begin{tabular}{cccccccc}
\hline
\multicolumn{1}{c}{\multirow{2}{*}{Synthetic   Dataset}} & \multirow{2}{*}{Pretraining} & \multirow{2}{*}{Training} & \multirow{2}{*}{Inference} & \multirow{2}{*}{Model Size} & \multicolumn{3}{c}{Rendering Quality} \\ \cline{6-8} 
\multicolumn{1}{c}{}                                     &                              &  &    &    & PSNR$\uparrow$  & SSIM$\uparrow$   & LPIPS$\downarrow$   \\ \hline
NeRF~\cite{mildenhall2020nerf}   & None     & 20 h    & 1/12 fps     & 14 MB                        & 31.0 dB             & 0.947    & 0.081     \\ \hline
IBRNet~\cite{wang2021ibrnet}                                                   & 1 day                        & 30 min                          & 1/25 fps                         & 15 MB                        & 28.1 dB             & 0.942    & 0.072     \\ \hline
MVSNeRF~\cite{chen2021mvsnerf}                                                  & 20 h                         & 15 min                          & 1/14 fps                         & 14 MB                        & 27.0 dB             & 0.931    & 0.168     \\ \hline
Plenoxels~\cite{yu2021plenoxels}                                          & None                         & 11 min                          & 15 fps                            & 1.1 GB                       & \textbf{31.7 dB}    & \textbf{0.958}    & \textbf{0.050}     \\ \hline
\rev{Plenoxels\_s~\cite{yu2021plenoxels}}   & \rev{None}      & \rev{8.5 min}     & \rev{18 fps}     & \rev{234 MB}   & \rev{28.5 dB}   & \rev{0.926}    & \rev{0.100}     \\ \hline
\rev{Pulsar~\cite{lassner2021pulsar}}   & \rev{None}       &   \rev{95 min}     &    \rev{4 fps}    &   \rev{228 MB}  &  \rev{26.9 dB}  &   \rev{0.923}   &  \rev{0.184}     \\ \hline
\rev{PBNR~\cite{kopanas2021point}}   & \rev{None}       & \rev{3 h}     & \rev{4 fps}     & \rev{2.96 GB}   &    \rev{27.4 dB} & \rev{0.932}  & \rev{0.164}       \\ \hline
Ours                                                     & \textbf{None}                & \textbf{3 min}                  & \textbf{32 fps}                   & \textbf{9 MB}    & 30.3 dB   & 0.945   & 0.078     \\ \hline
\end{tabular}

\label{tab:synthetic_result}

\end{table*}
\begin{figure*}[t!]
  \centering
   \includegraphics[width=0.95\linewidth]{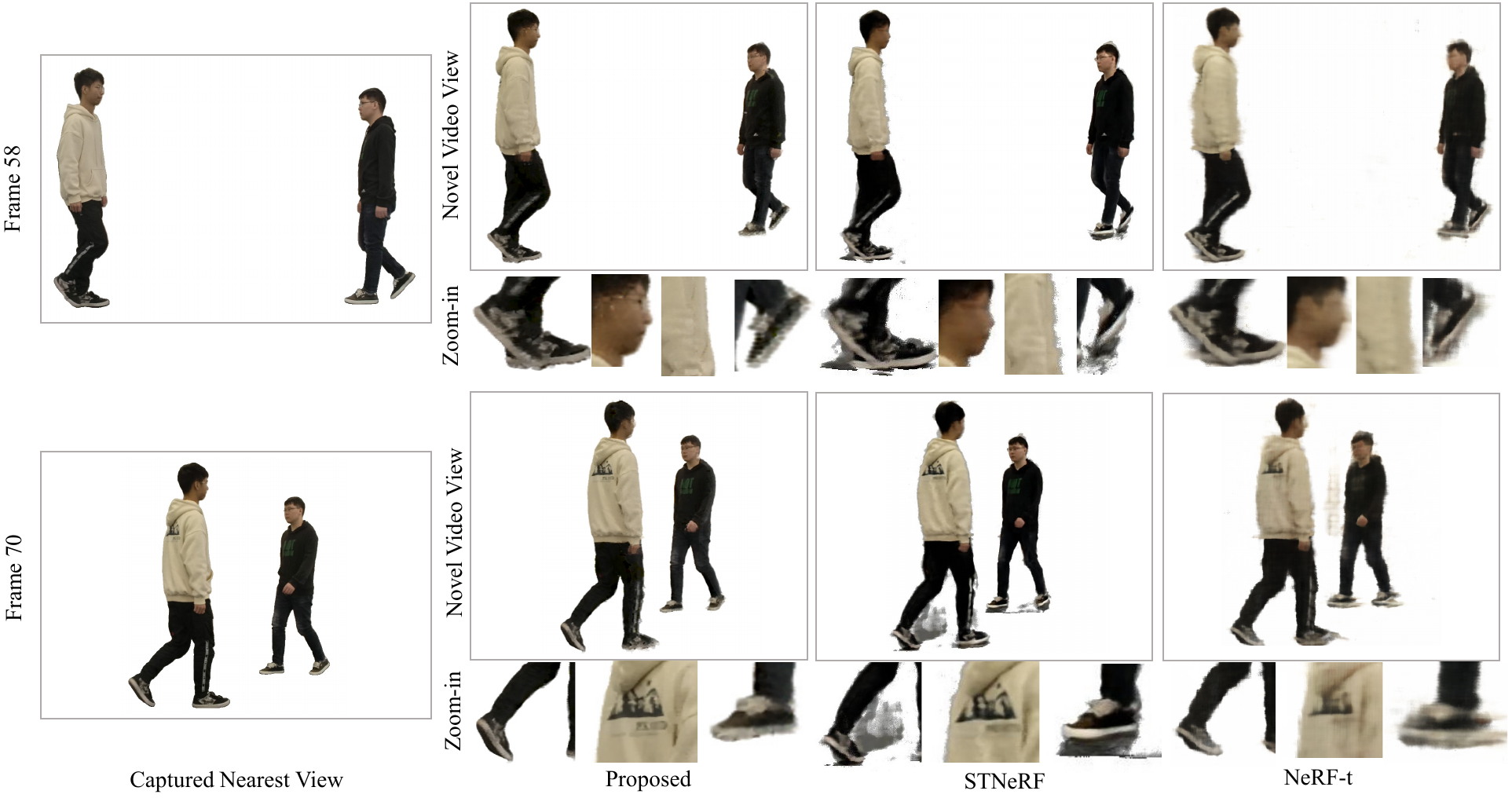}
   \vspace*{-8pt}
   \caption{\textbf{Novel Video View Synthesis.} We synthesize novel views for video timestamps of the STNeRF~\shortcite{zhang2021editable} dataset (Frame 58 and Frame70). Compared to STNeRF~\shortcite{zhang2021editable} and the NeRF-t model proposed in the same work, the differential point cloud radiance field approach produces substantially cleaner object boundaries, finer object detail and fewer reconstruction artifacts, especially in the presence of significant object motion, as illustrated in the insets. 
   }
   \label{fig:stnerf}
   \vspace{-3mm}
\end{figure*}

\vspace{-1mm}
\subsection{Video Synthesis}
When extending our method from novel image synthesis to novel video synthesis, we train a model per frame. This means that the training time grows as the frame number grows. However, in dynamic video scenes, nearby frames are temporally correlated. As such, we use the model learned in a given frame to initialize the next frame, thereby reducing the training time needed til convergence. 
For a given frame $1$, the proposed method obtains an accurate point cloud $S_1$ with learned spherical harmonics parameters. Next, we compute an initial point cloud $S_2$ for the next frame $2$, using the same scheme as discussed in Sec.~\ref{sec:training}. We then compute the Chamfer distance $\mathcal{CD}(S_1,S_2)$ between these two point clouds, that is
\begin{equation}
    \mathcal{CD}(S_1,S_2) = \frac{1}{S_1}\sum_{x \in S_1} \min_{y \in S_2} {||x-y||_2^2}+\frac{1}{S_2}\sum_{y \in S_2} \min_{x \in S_1} {||x-y||_2^2}.
\end{equation}
By minimizing the Chamfer distance via gradient descent (the Adam optimizer) applied to the point positions in $S_2$, we can obtain a more accurate initial point cloud for this next frame $2$, in terms of geometry. After refining the positions $p_i$ in $S_2$, we can compute the top-k closest points in $S_1$ and choose the average of their corresponding spherical harmonics parameters as the initial radiance field parameter of $p_i$. This strategy allows us to achieve a 4$\times$ improvement in training runtime.
\vspace{-1mm}

%% file: 5_results.tex
We evaluate the proposed point-based radiance representation for the novel view synthesis tasks in two scenarios: static scenes and dynamic video scenes. In both scenarios, we assume that measurements from multiple views are available as input, either as multi-view images or as videos from multiple views. In this section, we validate that the proposed method outperforms existing methods in computational efficiency while allowing for rendering high-fidelity images and videos from novel views. In addition, we provide extensive ablations that validate architecture choices we made for the proposed method. We implemented the proposed method in PyTorch, see Supplementary Code. We use a single NVIDIA 2080Ti GPU for training in all experiments. 

\vspace{-2mm}
\subsection{Novel View Synthesis for Static Scenes}
We test the proposed method on the rendered multi-view images from the \textit{Blender} dataset used in NeRF~\cite{mildenhall2020nerf}. \finalrev{Note that this dataset provides background masks that are automatically extracted from the Z-buffer during rendering}.
We compare our method against state-of-the-art  methods for novel-view synthesis: IBRNet~\cite{wang2021ibrnet}, MVSNeRF~\cite{chen2021mvsnerf}, \rev{Plenoxels~\cite{yu2021plenoxels}, Pulsar~\cite{lassner2021pulsar}, and PBNR~\cite{kopanas2021point}.
We note that PBNR requires pre-processed geometry information such as point clouds and point features retrieved from conventional SfM. In contrast, the proposed method is \emph{network-free} and learns both point cloud positions and appearance, without the need for separate input geometry. Quantitatively, although the full version of Plenoxel (model size of 1.1GB) performs better in terms of reconstruction quality than ours, its smaller version Plenoxel\_s (model size of 234MB) is 1.8dB worse than ours in PSNR, while our model size is only 9MB.} Table~\ref{tab:synthetic_result} shows that the proposed method outperforms existing methods in training speed, inference speed, and model size while maintaining comparable image quality. Figure~\ref{fig:synthetic} provides visualizations of the rendered images which validate that the synthesized images are on par with state-of-the-art methods. 


\vspace{-2mm}
\subsection{Video View Synthesis}

\rev{The computational efficiency and low memory footprint of the proposed point-based representation, allow it to explicitly represent challenging dynamic scenes better than implicit methods that rely on networks to produce time-conditional outputs as ST-NeRF and NeRF-T.} Note that such NeRF-based per-frame modeling of videos results in a heavy computational load: for example, a 1000-frame video corresponding to a 40-seconds clip at 25FPS requires more than 20K GPU hours (per frame training) to train for a vanilla NeRF model and more than 1 terabyte storage (when trained per frame) for a Plenoxels~\cite{yu2021plenoxels} model. Recently, Zhang et al.~\shortcite{zhang2021editable} proposed \finalrev{NeRF-T} and STNeRF as variants of NeRF that allow reducing the computational requirements of modeling videos for view synthesis by conditioning the radiance MLP of NeRF on the time $t$. While effective in reducing memory requirements, this comes at the cost of rendering quality compared to frame-by-frame modeling. The proposed method is light-weight and it allows for frame-by-frame learning with respect to both training time and storage. We evaluate the method on the video view synthesis datasets from STNeRF~\cite{zhang2021editable} and the human motion dataset DSC~\cite{vlasic2009dynamic}. \finalrev{Both of these two datasets contain background masks that are generated by a segmentation network and color thresholding, respectively}. \rev{We note that the authors of STNeRF released only a portion of their entire dataset, which we use to perform our experiments.} The dataset features 16 camera views distributed uniformly on a semicircle, from which we use 13 \rev{views} for training and 3 \rev{views} for testing. Table~\ref{tab:stnerf_result} and Figure~\ref{fig:stnerf} validate that our method outperforms the existing methods quantitatively and qualitatively. 
\rev{For experiments on DSC, please refer to the Appendix.}

\begin{table}[]
\caption{\textbf{Quantitative Evaluation on the STNeRF Dataset.} 
Compared to STNeRF and the NeRF variants suggested in \cite{zhang2021editable}, the training speed and inference speed of the proposed method is two orders of magnitude \rev{higher}. }
\vspace{-3mm}
\setlength{\tabcolsep}{1.4mm}{
\begin{tabular}{ccccccc}
\hline
\multicolumn{1}{c}{\multirow{2}{*}{STNeRF}} & \multirow{2}{*}{Train} & \multirow{2}{*}{Render} & \multirow{2}{*}{Model} & \multicolumn{3}{c}{Rendering Quality} \\ 
\multicolumn{1}{c}{}    &   &  &    & PSNR$\uparrow$  & SSIM$\uparrow$   & LPIPS$\downarrow$   \\ \hline
NeRF  & 40 h & 1/25 fps    & 14 MB  & 23.7 dB  & 0.853   & 0.304 \\ \hline
NeRF-t & 100 h & 1/26 fps  & 16 MB & 28.9 dB  & 0.913   & 0.259   \\ \hline
STNeRF   & 50 h  & 1/30 fps  & \textbf{12 MB} & 32.1 dB  & 0.918  & 0.224 \\ \hline
Ours  & \textbf{30 min} & \textbf{25 fps}  & 110 MB  & \textbf{34.6 dB}  & \textbf{0.927}    & \textbf{0.207}     \\ \hline
\end{tabular}}
\label{tab:stnerf_result}
\end{table}

\subsection{Ablation Experiments} \label{sec:ablation}
\rev{We evaluate each component of our method and validate their effectiveness.
First, we replace the spherical harmonics with learnable directionally invariant RGB values denoted as ``w/o sh''.
Second, we remove the hybrid coarse-to-fine strategy and only train the point cloud in a single stage denoted as ``w/o hybrid''. 
Third, we remove the consistency filter function \finalrev{(refer to the Appendix for additional details on initialization)} for every epoch during training time, noted as "w/o filter".
}
Table~\ref{tab:ablation_module} shows that using spherical harmonics, the coarse-to-fine strategy, and the consistency filter function jointly achieves the best view synthesis performance. \rev{Please refer to the Appendix for corresponding qualitative results and additional ablation experiments.}

\begin{table}[]
\vspace{-3mm}
\caption{\textbf{Model Ablation Experiments}. We evaluate the rendering quality of our method \rev{on the Blender dataset~\shortcite{mildenhall2020nerf}} when gradually removing components from the rendering pipeline. Specifically, we ablate the spherical harmonics model per point, the coarse-to-fine strategy, and the filtering function for the training. The experimental results validate that all components contribute to the rendering quality.}
\vspace{-3mm}
\begin{tabular}{lccc}
\hline
\multicolumn{1}{c}{\multirow{2}{*}{Module}} & \multicolumn{3}{c}{Rendering Quality}      \\
\multicolumn{1}{c}{}                        & PSNR$\uparrow$  & SSIM$\uparrow$   & LPIPS$\downarrow$       \\ \hline
w/o sh       & \rev{27.1 dB}    & \rev{0.932} & \rev{0.155}      \\ \hline
w/o hybrid   & \rev{26.6 dB}    & \rev{0.925}  & \rev{0.163}          \\ \hline
w/o filter  & \rev{29.1 dB}    & \rev{0.941} & \rev{0.103}      \\ \hline
Ours   & \rev{\textbf{30.3 dB}} & \rev{\textbf{0.945}} & \rev{\textbf{0.078}} \\ \hline
\end{tabular}
\vspace{-1mm}
\label{tab:ablation_module}
\end{table}

%% file: 8_conclusion.tex
Departing from volumetric representations of scene radiance, extensively explored in recent neural rendering work, we investigate an alternate radiance field representation that is based on points without giving up end-to-end differentiability. In contrast to coordinate-network representations that require ray-marching with hundreds of network evaluations per ray, the proposed representation allows us to render a novel view via differentiable splat-based rasterization. As a result, the proposed method is two orders of magnitude faster in training and inference time than volumetric methods, such as NeRF, while being on par in quality. The method does not require any pre-training and the representation is highly compact, occupying only 10~Mb per static scene. Future directions that may build on the approach introduced in this work could include hybrid point/surface models, or combine the proposed rendering method with point-based physical simulation methods.

%% file: appendix.tex

\finalrev{
\subsection*{Additional Evaluations on the Synthetic Dataset}
}

Tab.\ref{tab:nerf_details} reports PSNR for every object for the static novel view synthesis experiment. Our method achieves quantitative results comparable to Plenoxels~\cite{yu2021plenoxels} and NeRF~\cite{mildenhall2020nerf} while being resource-efficient in terms of memory and train/test runtime.

\begin{table}[h]
\caption{\rev{\textbf{Additional Static View Synthesis Evaluation.} The proposed method achieves comparable performance to existing state-of-the-art view synthesis methods for static scenes while being more efficient in training/inference speed and memory usage.}}
\label{tab:nerf_details}
{ \resizebox{\columnwidth}{!}{
 \rev{
\begin{tabular}{lccccccccc}
\hline
          & Chair & Drums & Ficus & Hotdog & Lego  & Materials & Mic   & Ship  & Avg.   \\ \hline
NeRF      & 33.00 & 25.01 & 30.13 & 36.18  & 32.54 & \textbf{29.62}     & 32.91 & 28.65 & 31.01 \\ \hline
Plenoxels & \textbf{33.98} & 25.35 & \textbf{31.83} & \textbf{36.43}  & \textbf{34.10} & 29.14     & 33.26 & \textbf{29.62} & \textbf{31.71} \\ \hline
Ours      & 32.98 & \textbf{25.53} & 29.01 & 34.56  & 31.33 & 28.01     & \textbf{33.82} & 27.01 & 30.28 \\ \hline
\end{tabular}}
}}
\end{table}
%
\finalrev{
\subsection*{Discussion on the Usage of Background Masks}
One limitation of our method is that it requires background masks to produce the initial point cloud. Although the improvements in training and testing time are not results of this initialization but the point cloud representation itself, it is an important step of the proposed approach. In our experiments, we consider datasets that provide the masks for single-object scenes, either generated by a separate segmentation network or extracted from the Z-Buffer for synthetic scenes. For more complex scenes which are more challenging to segment future work could extend the proposed method by relaxing the consistency check function to a soft check. 
}
\finalrev{Next, we discuss the usage of masks in the competing methods from Tab.\ref{tab:nerf_details} above. Using masks in NeRF is investigated in VaxNeRF~\shortcite{kondo2021vaxnerf}. In reconstruction quality, the proposed method fares on average 0.1~dB above the results of VaxNeRF. During inference, the proposed method (32 fps) is two orders of magnitude faster than VaxNeRF (0.1 fps). As such, the proposed model not only benefits from masks, but also from the representation and optimization method. To train a Plenoxel model, Yu et al.~\shortcite{yu2021plenoxels} require masks to determine the bounding volume to optimize for a Plenoxel model. Finally, we note that methods like Pulsar~\shortcite{lassner2021pulsar} require more than just masks. Pulsar uses conventional SfM (COLMAP) to build a dense point cloud, which it then passes to SIBR to get a depth map and a normal map. We note that for single objects, a rich body of work has explored background extraction methods using such depth maps.}

\begin{figure}[h]
  \centering
  \includegraphics[width=0.92\linewidth]{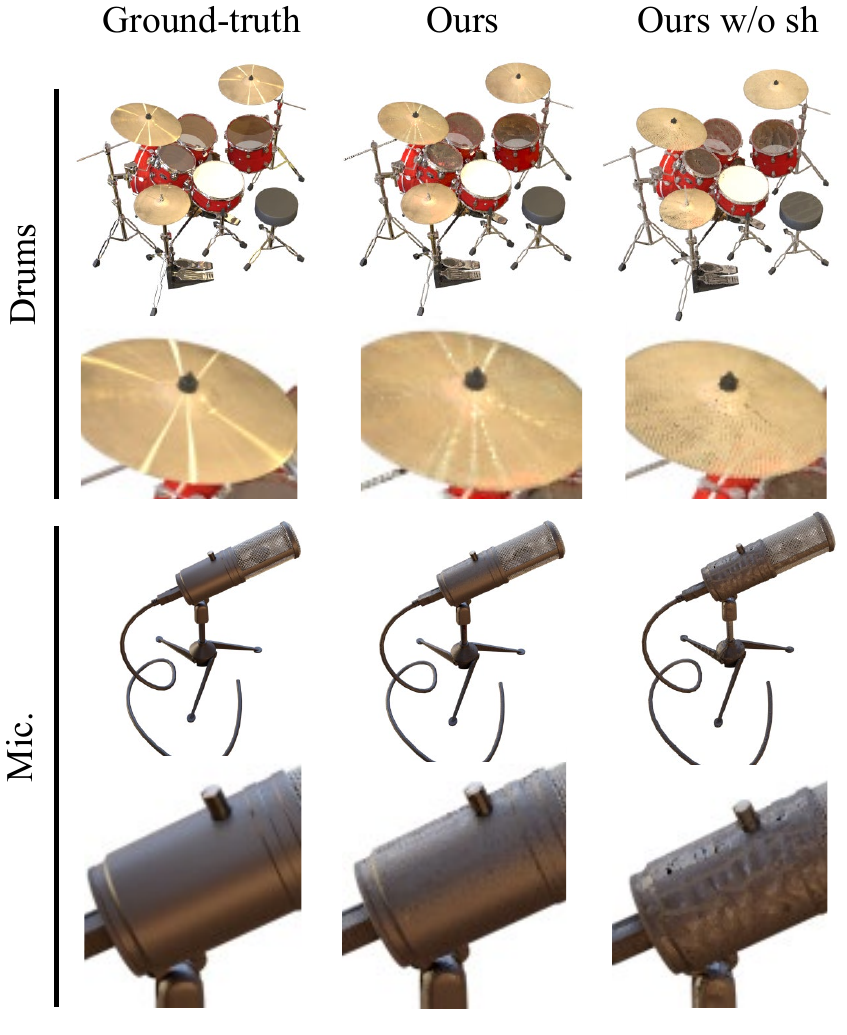}
  \caption{\rev{\textbf{Ablating View Dependence.} We visualize the ablations of the learnable spherical harmonics. For the mic object, we observe noticeable artifacts on the object surface if we remove the view-dependence module. This is due to the optimization with color inconsistency from different views.}}
  \label{fig:shablation}
\end{figure}

\begin{figure}[h]
  \centering
   \includegraphics[width=0.98\linewidth]{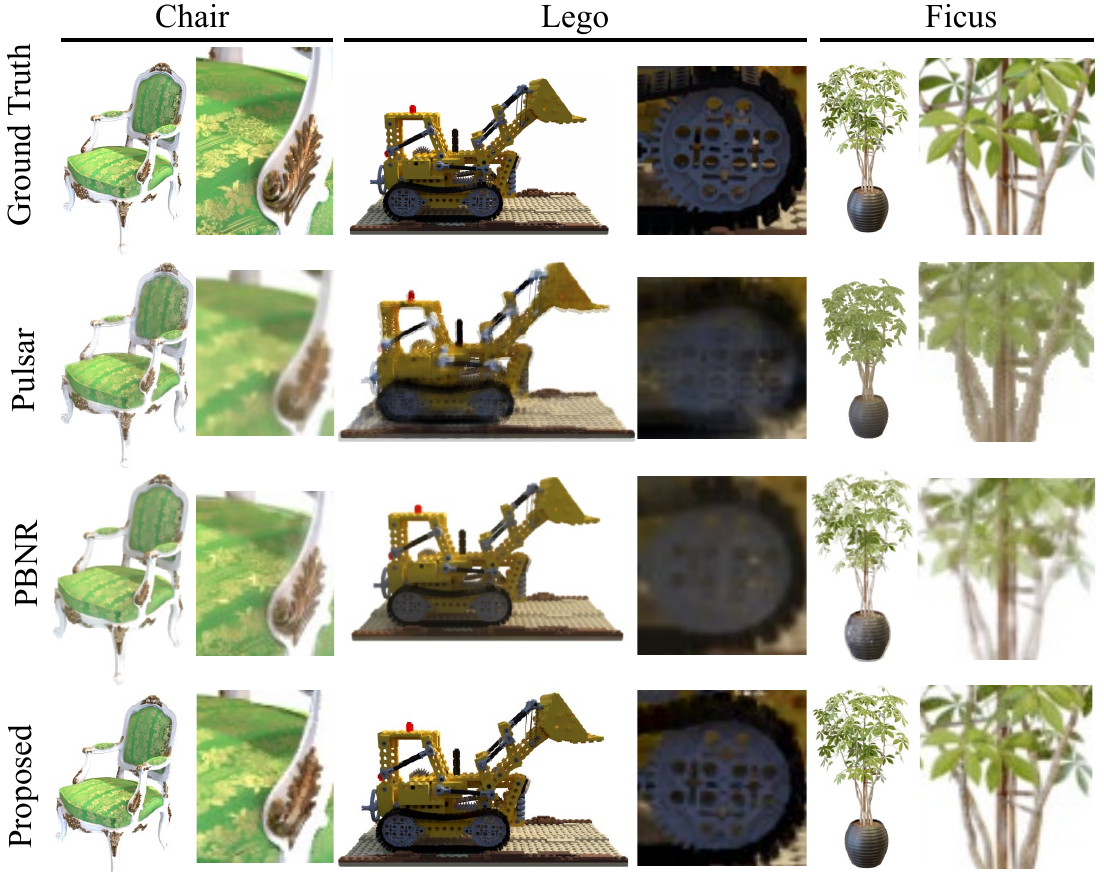}
   \caption{\rev{\textbf{Additional Comparison to Point-based Methods.} We visualize novel views rendered for the Chair, Lego and Ficus objects from the Blender dataset from~\cite{mildenhall2020nerf}, validating that our method outperforms existing point-based view synthesis methods.}}
   \label{fig:more_cmp}
\end{figure}

\rev{\subsection*{Additional Visualization on the Blender Dataset}
Fig.\ref{fig:more_cmp} provides further visualizations for results on Blender dataset from~\cite{mildenhall2020nerf}. The proposed approach achieves substantially better rendering quality than the two recent point-based rendering baselines: Pulsar and PBNR.
}

\begin{figure}
  \centering
  \includegraphics[width=0.8\linewidth]{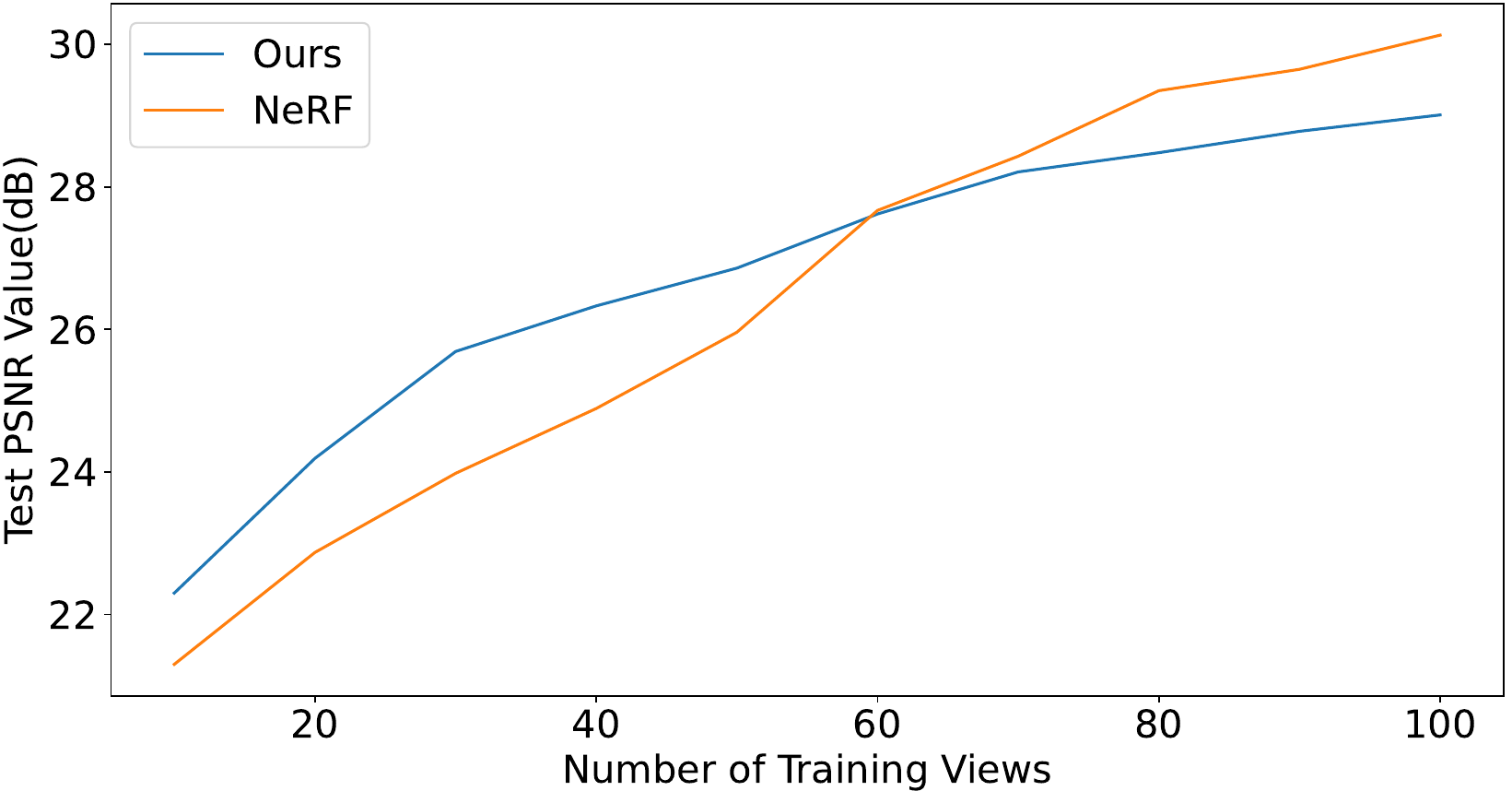}
  \caption{\textbf{View Ablations.} We evaluate the quality of the proposed point-based method and the volumetric NeRF~\shortcite{mildenhall2020nerf} on the 'Ficus' scene~\shortcite{mildenhall2020nerf} with a gradually reduced number of training views. The proposed sparse representation allows for reconstructions with reasonable quality even with as few as 20 views, where NeRF drops by more than 10~dB in PSNR.}
  \label{fig:sparse}
\end{figure}

\begin{figure}[h]
  \centering
  \includegraphics[width=0.92\linewidth]{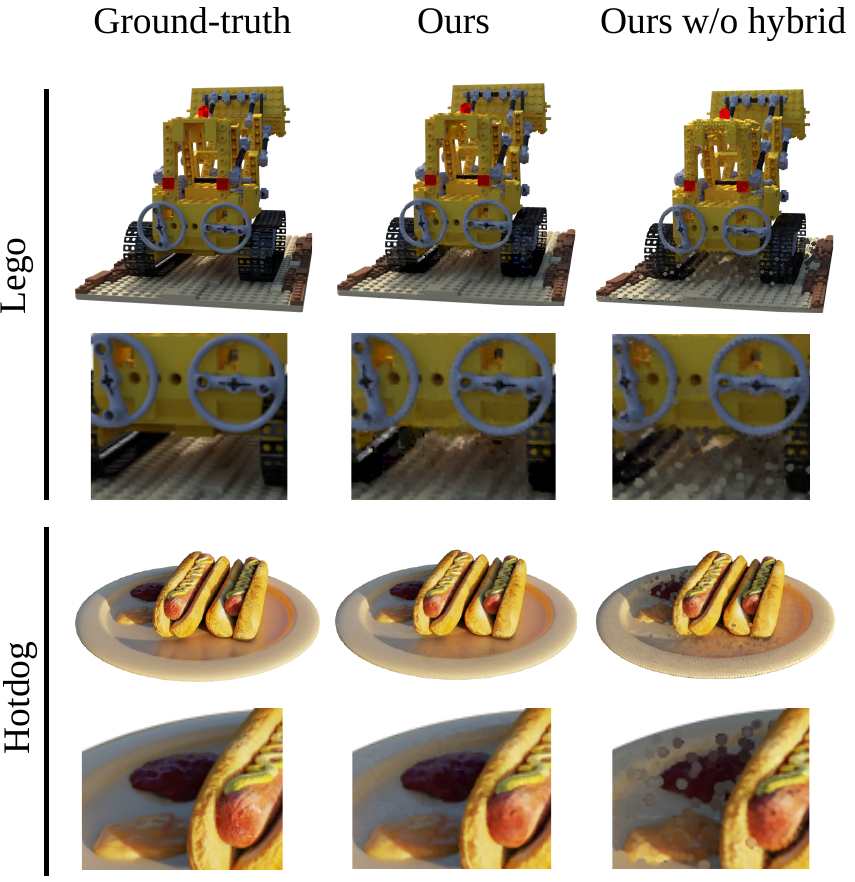}
  \caption{\rev{\textbf{Analysis of Coarse-to-Fine Optimization.} We analyze rendering quality with and without the hybrid coarse-to-fine training strategy. We observe noticeable outlier points if we train the model without the coarse-to-fine training strategy, validating the proposed approach.}}
  \label{fig:hybridablation}
\end{figure}

\begin{figure*}[h]
  \centering
   \includegraphics[width=0.98\linewidth]{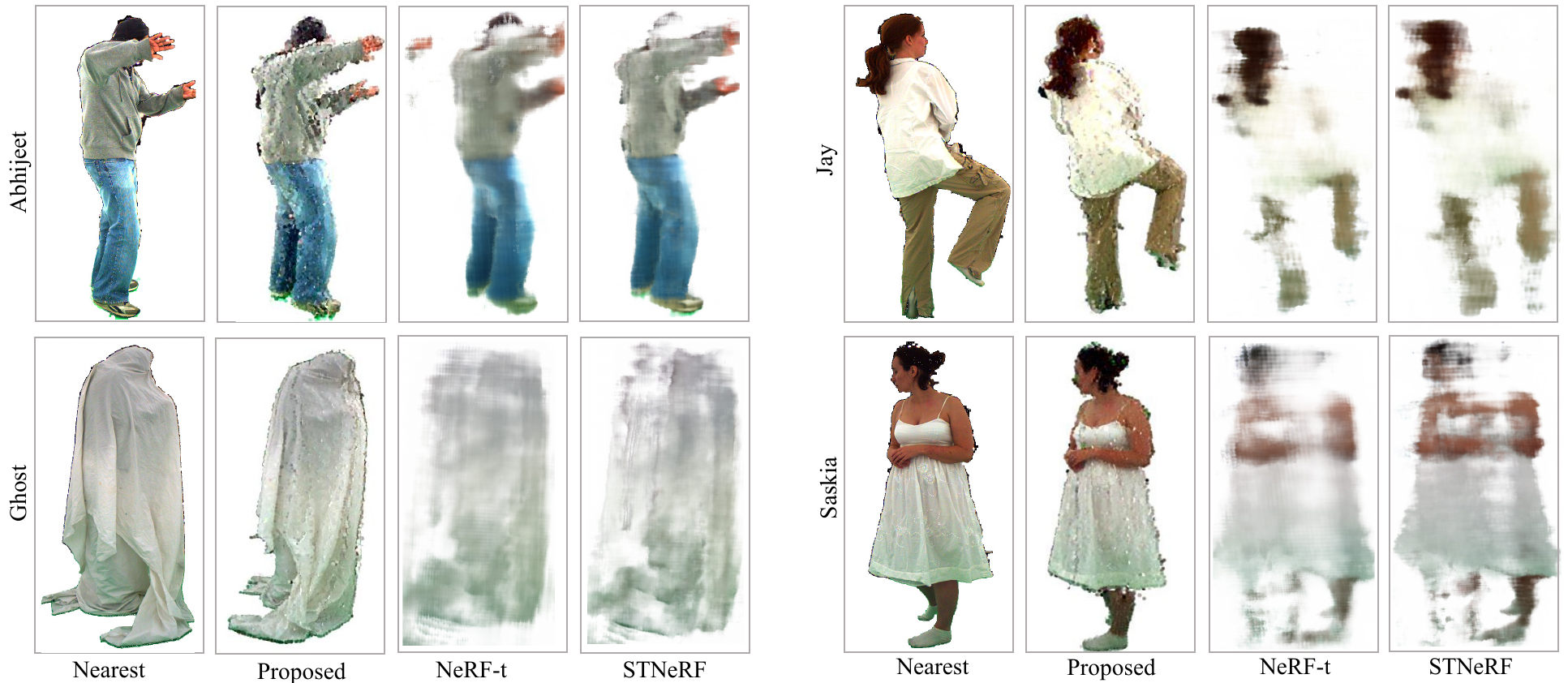}
  \vspace{3mm}
   \caption{\textbf{Novel View Synthesis on the DSC Dataset.} We synthesize novel views for videos from the DSC dataset\shortcite{vlasic2009dynamic}  (frame 50 for every scene). We show the nearest captured frame on the left. We retrained STNeRF~\shortcite{zhang2021editable} and the NeRF-t variant suggested by the authors on the same data. In contrast to the proposed method, these existing approaches fail in highly dynamic scenes (right).}
   \label{fig:dsc_visual}
   \vspace{3mm}
\end{figure*}

\rev{
\subsection*{Additional Ablation Studies}
We evaluate the impact of the number of points on the final rendering quality to assess the trade-off between reconstruction accuracy and computational efficiency, reported Table~\ref{tab:ablation_pointnumber}. While a large number of points can better represent a scene radiance field, at the cost of higher computing and memory, diminishing returns can be observed at circa 50k points for the given static scenes we evaluate the method on. 

We also assess the proposed method with and without modeling view dependence with spherical harmonics in Fig.\ref{fig:shablation}. Modeling view dependence is essential for the proposed model to represent specular highlights, as apparent on the brass symbols and the plastic surface on the microphone object. These experiments further validate the proposed model that relies on learnable spherical harmonics.}

\finalrev{In Figure~\ref{fig:sparse}, we also provide an ablation study for sparse training views on the ``Ficus'' object from the Synthetic Blender dataset. Although our method performs slightly worse than the NeRF baseline when training on the whole 100 views, the method gradually improves on NeRF when the training images become sparser.}

\subsection*{\rev{Additional Evaluations on the DSC Dataset}}
\rev{In the DSC dataset, 8 cameras are distributed in a 360$^\circ$ setup and we use all of them to train the model (due to the overall low number of camera views). We argue that view synthesis on DSC is very challenging since there are only 8 training views available, \finalrev{covering 360 degrees, and with slight pose miscalibration.} We report quantitative results in Table~\ref{tab:dsc_result} and qualitative results in Figure~\ref{fig:dsc_visual} that further corroborate the findings we made for the STNeRF dataset.  In all cases, the proposed method achieves high-quality view synthesis results with orders of magnitude decreased training time and improved inference speed.}

\begin{figure}[h]
  \centering
  \includegraphics[width=0.92\linewidth]{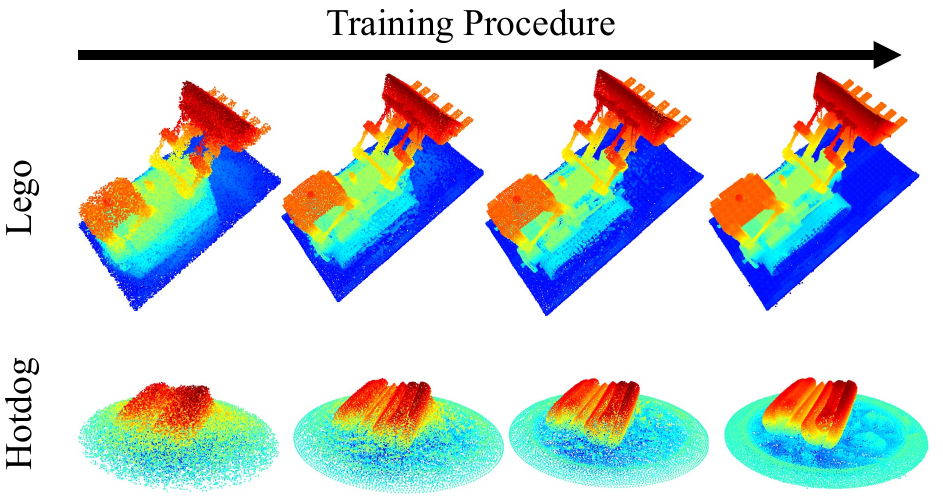}
  \caption{\rev{\textbf{Point Cloud Visualizations Evolving Throughout Training.} As training progresses, the learned point cloud approximates the object geometry more accurately, illustrating the effectiveness of the positional updates of the proposed optimization method. }}
  \label{fig:ptsevolution}
\end{figure}

\begin{table}[h]
\caption{\textbf{Quantitative Evaluation on the DSC Dataset.} We report the training time, rendering speed and the model size required to represent scenes from the DSC\cite{vlasic2009dynamic} dataset with the proposed method. Note that we train a radiance point cloud for each frame here and we report the total memory consumption.}
\begin{tabular}{ccccccc}
\hline
\multicolumn{1}{c}{\multirow{2}{*}{STNeRF}} & \multirow{2}{*}{Train} & \multirow{2}{*}{Render} & \multirow{2}{*}{Model}  \\ 
\multicolumn{1}{c}{}    &   &  &      \\ \hline
NeRF  & 50 h & 1/15 fps    & 14 MB   \\ \hline
NeRF-t & 75 h & 1/15 fps  & 16 MB   \\ \hline
STNeRF   & 86 h  & 1/18 fps  & \textbf{12 MB}  \\ \hline
Ours  & \textbf{1 h} & \textbf{28 fps}  & 240 MB     \\ \hline
\end{tabular}
\vspace{1mm}
\label{tab:dsc_result}
\end{table}

\begin{table}[h]
\caption{\rev{\textbf{Point Cloud Size Ablations}. The rendering quality of the proposed method grows with the density of the point cloud (for a given volume), at the cost of training time, inference speed and model size. We trade off these dimensions and choose 45k points for our experiments.}}.
\label{tab:ablation_pointnumber}
\rev{
\begin{tabular}{ccccc}
\hline
\multirow{2}{*}{Point Scale} & \multirow{2}{*}{Train} & \multirow{2}{*}{Render} & \multirow{2}{*}{Model} & \multirow{2}{*}{PSNR$\uparrow$} \\ &  &  &     &      \\\hline
15 k   & 16 min   & 35 fps    & 37 MB      & 32.2 dB    \\ \hline
30 k   & 20 min   & 32 fps    & 73 MB      & 33.6 dB    \\ \hline
\textbf{45 k} & 30 min & 25 fps     & 110 MB       & 34.6 dB   \\ \hline
60 k  & 41 min  & 16 fps  & 146 MB & 34.8 dB     \\ \hline
100 k & 76 min &	10 fps &	244 MB &	34.2 dB  \\ \hline
\end{tabular}
}
\vspace{1mm}
\end{table}


\rev{
\subsection*{Additional Initialization Details}
Next, we provide further details on the point cloud initialization from initial image masks. We first uniformly sample points from scene volume. For each 3D point $P^j$, we compute the projected 2D point $p_i^j$ on a camera $i$ using Eq.\ref{eq:projection}. After each projection, we check whether the corresponding pixel is a background pixel, in which case we discard the point. Specifically, we define the following consistency check function
$$
\mathcal{CC}(p^j) = \mathbb I(\sum_{i=1}^N \mathbb I (M_i(p_i^j)>0)=N),
$$
where $M_i(\cdot)$ is the foreground mask and $\mathbb I(\cdot)$ is the indicator function. We compute the value of this consistency function for every sample point and only keep the point if the value is true. We also use this consistency function during the training time by filtering out points that become inconsistent after every training epoch. 

}

\begin{algorithm}
	\rev{\caption{\rev{Hybrid Coarse-to-Fine Update Strategy}\label{alg:hybrid}} }
	\begin{algorithmic}[1]
		\REQUIRE Coarse point cloud $S=\{P_i|P_i=(x_i,y_i,z_i,f_i)\}$, 
		\REQUIRE Hyper-parameters $N,K,\epsilon$ 
		\ENSURE New point cloud $S^{'}=\{P_i|P_i=(x_i,y_i,z_i,f_i)\}$ 
		\STATE $S^{'} = \emptyset$, \quad Partition the space into $N\times N \times N$ voxels. \\
		\FOR {each voxel}
		\STATE Fetch all the points $\{P_1,P_2,...,P_M\}$ inside this voxel.
		\STATE Compute the average: $P_a=\frac{1}{M}\sum\limits_{i=1}^M$.
		\STATE Add $P_a$ to $S^{'}$.
		\ENDFOR
		
		\FOR {$P_i \in S^{'}$}
		\STATE Fetch neighbor $K$ points $\{P_1,P_2,...,P_K\}$ for $P_i$.
		\STATE Compute distances between neighbors and $P_i$ to get $\{d_1,d_2,...,d_K\}$
		\STATE Compute standard deviation $\delta$ for set $\{d_1,d_2,...,d_K\}$
    		\IF{$\delta > \epsilon$} 
    		\STATE Remove $P_i$ from $S^{'}$
    		\ENDIF
		\ENDFOR
		
		\FOR {$P_i \in S^{'}$}
		\STATE Fetch neighbor $K$ points $\{P_1,P_2,...,P_K\}$ for $P_i$.
		\STATE Compute the average: $P_a=\frac{1}{K}\sum\limits_{i=1}^K P_i$.
		\STATE Add $P_a$ to $S^{'}$.
		\ENDFOR
	\RETURN $S^{'}$
	\end{algorithmic} 
\end{algorithm}

\subsection*{Additional Algorithm Details}
To supplement the description in the main text, Algorithm~\ref{alg:hybrid} contains pseudo-code of our coarse-to-fine update method. We further note that all code, models and data required to reproduce the findings in this work have been published.

\subsection*{Point Cloud Evolution During Training}
We visualize the evolution of the learned point clouds during training in Fig.\ref{fig:ptsevolution}. As the training progresses, we can observe three trends: the point cloud approximates the target geometry more accurately with more training, the outliers are eliminated with additional training stages, and the concave regions are progressively improved with additional training iterations. These additional visualization qualitatively validate the proposed training scheme.
